\documentclass[11pt]{article}
\usepackage[]{emnlp2021}
\usepackage{times}
\usepackage{latexsym}

\pdfoutput=1

\usepackage{graphicx}
\usepackage{subfigure}
\usepackage{tabularx}
\usepackage{courier}
\usepackage{xcolor,colortbl}
\usepackage{multirow,tabularx}
\RequirePackage{amsmath,amssymb,bm}
\RequirePackage{enumitem}

\usepackage{microtype}

\def\aclpaperid{2418} 

\usepackage[T1]{fontenc}

\usepackage[utf8]{inputenc}
\usepackage[textsize=scriptsize]{todonotes}
\RequirePackage{xspace}

\usepackage{microtype}
\usepackage{makecell}
\usepackage[linesnumbered,ruled,vlined,norelsize,algo2e]{algorithm2e}
\usepackage{multirow}
\newcommand{\base}{TaBERT+MAPO}
\newcommand{\baseqg}{TaBERT+MAPO+QG}

\newcommand{\tbert}{$\text{TaBERT}_{t}$+MAPO}
\newcommand{\tbertqg}{$\text{TaBERT}_{t}$+MAPO+QG}

\newcommand{\ws}{Wiki\-SQL-TS\xspace}
\newcommand{\wtq}{Wiki\-TQ-TS\xspace}
\newcommand{\sys}{\textsc{T3QA\ }}
\usepackage{color}

\newcommand{\eat}[1]{} 

\eat{}

\def\ztitle{Topic Transferable Table Question Answering}
\title{\ztitle}

\author{
Saneem A. Chemmengath$^{1*}$ \quad
Vishwajeet Kumar$^{1}$\thanks{~Equal contribution by first two authors.}\quad
Samarth Bharadwaj$^{1}$\\
\textbf{Jaydeep Sen$^{1}$ \quad
Mustafa Canim$^{1}$,\quad
Soumen Chakrabarti$^{2}$} \\
\textbf{Alfio Gliozzo$^{1}$ \quad
Karthik Sankaranarayanan$^{1}$,
} \\
$^{1}$IBM Research \quad
$^{2}$Indian Institute of Technology, Bombay
\\
\small\texttt{saneem.cg@in.ibm.com, vishk024@in.ibm.com, samarth.b@in.ibm.com,}\\
\small\texttt{jaydesen@in.ibm.com, mustafa@us.ibm.com, soumen.chakrabarti@gmail.com}\\
\small\texttt{gliozzo@us.ibm.com, kartsank@in.ibm.com}
\\
}

\begin{document}
\maketitle

\begin{abstract}
Weakly-supervised table question-answering (TableQA) models have achieved state-of-art performance by using pre-trained BERT transformer to jointly encoding a question and a table to produce structured query for the question. However, in practical settings TableQA systems are deployed over table corpora having topic and word distributions quite distinct from BERT's pretraining corpus. In this work we simulate the practical topic shift scenario by designing novel challenge benchmarks \ws{} and \wtq\footnote{The source code and new dataset splits are available at  \url{https://github.com/IBM/T3QA}}, consisting of train-dev-test splits in five distinct topic groups, based on the popular WikiSQL and Wiki\-Table\-Questions datasets.
We empirically show that, despite pre-training on large open-domain text, performance of models degrades significantly when they are evaluated on unseen topics. 
In response, we propose \sys (Topic Transferable Table Question Answering) a pragmatic adaptation framework for TableQA comprising of: (1)~topic-specific vocabulary injection into BERT, (2)~a novel text-to-text transformer generator (such as T5, GPT2) based natural language question generation pipeline focused on generating topic specific training data, and (3)~a logical form re-ranker.  We show that \sys provides a reasonably good baseline for our topic shift benchmarks. \eat{These three enhancements mitigate the performance drop caused by topic-shift on all our benchmark topics.} We believe our topic split benchmarks will lead to robust TableQA solutions that are better suited for practical deployment.

\end{abstract}


\section{Introduction}

Documents, particularly in enterprise settings, often contain valuable tabular information (e.g., financial, sales/marketing, HR). Natural language question answering systems over a table (or TableQA) have an additional complexity of understanding the tabular structure including row/column headers compared to the more widely-studied passage-based reading comprehension (RC) problem. Further, TableQA may involve complex questions with multi-cell or aggregate answers. 

Most of the TableQA systems use semantic parsing approaches that utilizes language encoders to produce an intermediate logical form from the natural language question which is executed that over the tabular data to get the answer. While some systems \cite{zhong2017seq2sql} were fully supervised, needing pairs of questions and logical forms as training data, more recent systems \cite{pasupat2015compositional, krishnamurthy-etal-2017-neural, dasigi-etal-2019-iterative}
rely only on the answer as weak supervision and search for a correct logical form. The current best TableQA systems \cite{herzig2020tapas, yin2020tabert} capitalize on advances in language modeling, such as BERT, and extend it to encode table representations as well. They are shown to produce excellent results on  popular benchmarks such as WikiSQL~\cite{zhong2017seq2sql} and WikiTableQuestions(WikiTQ)~\cite{pasupat2015compositional}. 

\begin{figure}[!t]
\small
\centering
\begin{tabular}{|l|l|r|}
\hline
\textbf{Party} & \textbf{Candidate} & \textbf{Votes} \\ \hline
Conservatives & Andrew Turner  & 32,717 \\ 
Liberal Democrats & Anthony Rowlands &  19,739\\ 
Labour & Mark Chiverton & 11,484 \\ 
UK Independence & Michael Tarrant & 2,352 \\ 
Independent & Edward Corby & 551 \\ \hline
\end{tabular}
\includegraphics[width=.9\hsize]{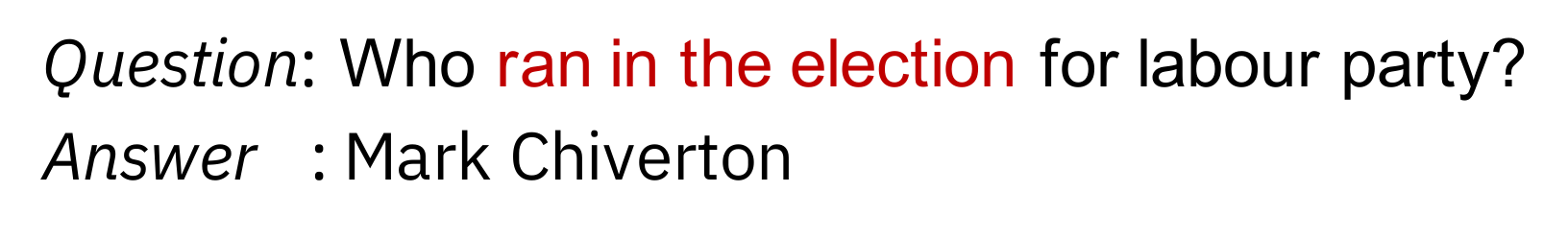}
\caption{Topic-sensitive representations are important to infer that, in the context of the topic \emph{politics}, the query span ``ran in the election'' should be linked to the ``Candidate'' column in the table.}
\label{fig:ex}
\end{figure}

\begin{figure}[!t]
\centering
\includegraphics[width=.48\textwidth]{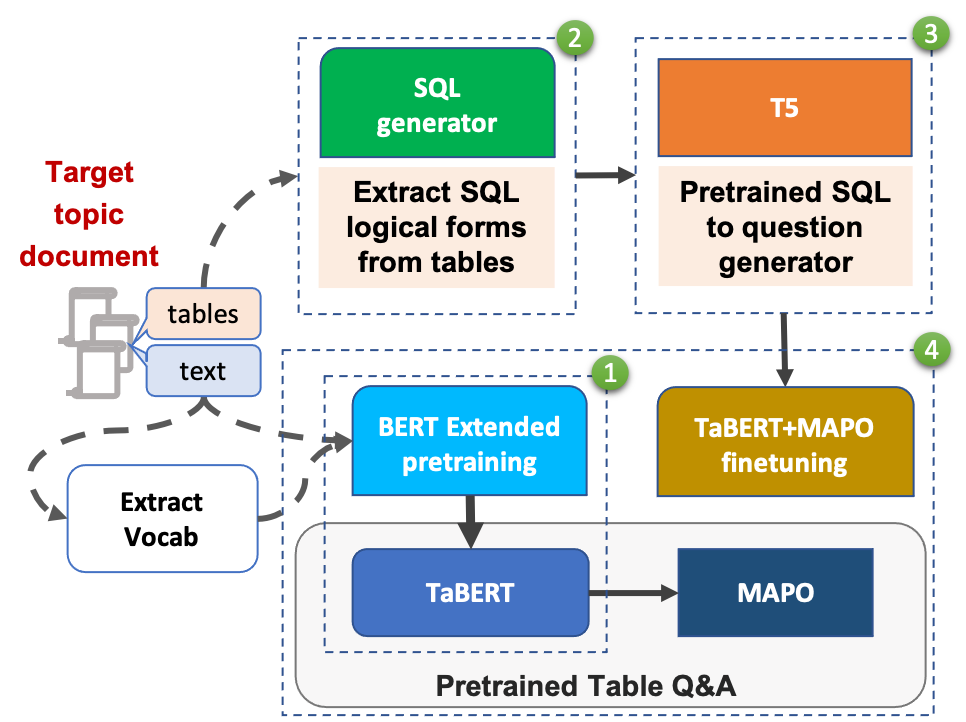}
\caption{Overview of the proposed T3QA framework for weakly-supervised TableQA.}
\label{fig:proposed}
\end{figure}

With increasing prevalence of text analytics as a centrally-trained service that serves diverse customers, practical QA systems will encounter tables and questions from topics which they may not have necessarily seen during training.  It is critical that the language understanding and parsing capabilities of these QA models arising from their training regime are sufficiently robust to answer questions over tables from such unseen topics.

As we show later in this paper, the existing approaches degrade significantly when exposed to questions from topics not seen during training (i.e., topic-shift).\footnote{Topic shift may be regarded as a case of domain shift studied in the ML community. However, here we refrain from referring to the proposed topic-driven splits as ``domains'' due to the open-domain nature of these datasets and the pre-training data used to build these models.}
To examine this phenomenon, we first instrument and dissect the performance of these recent systems under topic shift. In particular, we experiment with TaBERT~\cite{tabert}, which is a weakly supervised TableQA model
which encodes the table and question using BERT-encoder and outputs a logical form using an LSTM decoder.
In the example shown in Figure~\ref{fig:ex}, topic shift may cause poor generalization for specific terminology or token usage across unseen topics.

We introduce a novel experimental protocol to highlight the difficulties of topic shift in the context of two well-known Wikipedia-based TableQA datasets: WikiSQL~\cite{zhong2017seq2sql} and WikiTableQuestions~\citet{pasupat2015compositional}. Despite recent transformer-based TableQA models being pre-trained with open-domain data, including Wikipedia itself, we observe a performance drop of 5--6\% when test instances arise from topics not seen during training.

To address this challenge, we next propose a novel \sys framework for TableQA training that leads to greater cross-topic robustness. Our approach uses only unlabeled documents with tables from the never-seen topic (which we interchangeably call the \emph{target} topic), without any hand-created (question,logical form) pairs in the target topic. Specifically, we first extend the vocabulary of BERT for the new topic. Next, it uses a powerful text-to-text transfer transformer module to generate synthetic questions for the target topic. A pragmatic question generator first samples SQL queries of various types from the target topic table and transcribes them to natural language questions which is then used to finetune the TableQA model on target topic. Finally, \sys improves the performance of the TableQA model with a post-hoc logical form re-ranker, aided by entity linking. The proposed improvements are applicable to any semantic parsing style TableQA with transformer encoders and is shown to confer generally cumulative improvements in our experiments. To the best of our knowledge, this is the first paper to tackle the TableQA problem in such a zero-shot setting with respect to target topics.

The main contributions of this work are:
\begin{itemize}
    \item This is the first work to address the phenomenon of \textit{topic shift} in Table Question Answering systems.
    \item We create novel experimental protocol on 2 existing TableQA datasets to study the effects of topic shift. (\ws and \wtq) 
    \item We propose new methods that uses unlabeled text and tables from target topic to create TableQA models which are more robust to topic shift.
\end{itemize}

\section{Related work}
\eat{Our work revolves around three broad dimensions: (1) TableQA (2) Question Generation (QG) and (3) Domain Adaptation. We review the related works for each of these.
}
\begin{table*} [ht!]\footnotesize
\centering
\begin{tabular}{|l|l|c|c|c|c|c|c|} 
\hline
\multirow{2}{*}{\textbf{Topics} } & \multirow{2}{*}{\textbf{Member sub-topics from Wikipedia} } & \multicolumn{3}{c|}{\textbf{\ws} } & \multicolumn{3}{c|}{\textbf{\wtq}} \\ 
\cline{3-8}
 &  & \textbf{Train}  & \textbf{Dev}  & \textbf{Test}  & \textbf{Train}  & \textbf{Dev}  & \textbf{Test}  \\ 
\hline
Politics & \begin{tabular}[c]{@{}l@{}}Crime, Geography, Government, Law, Military, Policy, \\Politics, Society, World \end{tabular} & 7728 & 1236 & 2314 & 1836 & 545 & 580 \\ 
\hline
Culture & \begin{tabular}[c]{@{}l@{}}Entertainment, Events, History, Human \\behavior, Humanities, Life, Culture, Mass media, \\Music, Organizations \end{tabular} & 11804 & 1734 & 3198 & 2180 & 502 & 691 \\ 
\hline
Sports & Sports & 26090 & 4016 & 7242 & 4867 & 1195 & 1848 \\ 
\hline
People & People & 6548 & 861 & 1957 & 1946 & 420 & 743 \\ 
\hline
Misc & \begin{tabular}[c]{@{}l@{}}Academic disciplines, Business, Concepts, Economy, \\Education, Energy, Engineering, Food and Drink, Health, \\Industry, Knowledge, Language, Mathematics, Mind, \\Objects, Philosophy, Religion, Nature, \\Science and technology, Universe \end{tabular} & 3059 & 395 & 852 & 1032 & 357 & 438 \\
\hline
\end{tabular}
\caption{\label{tab:topics} Statistics of the proposed  \ws~and \wtq~benchmarks per topic.}
\end{table*}

Most TableQA systems take a semantic parsing view~\cite{pasupat2015compositional, zhong2017seq2sql,nsm} for question understanding and produce a logical form of the natural language question. Fully-supervised approaches, such as by \cite{zhong2017seq2sql} need pairs of questions and logical form for training.  However, obtaining logical form annotations for questions at scale is expensive. A simpler, cheaper alternative is to collect only question-answer pairs as weak supervision \cite{pasupat2015compositional, krishnamurthy-etal-2017-neural, dasigi-etal-2019-iterative}.  Such systems search for the correct logical forms under syntactic and semantic constraints that produce the correct answer. Weak supervision is challenging, owing to the large search space that includes many possible spurious logical forms \cite{guu2017language} that may produce the target answer but not an accurate logical transformation of the natural question. 

Recent TableQA systems \cite{herzig2020tapas, yin2020tabert, rci} extend BERT to encode the entire table including headers, rows and columns. They aim to learn a table-embedding representation that can capture correlations between question keywords and target cell of the table. \eat{However, encoding large tables in this way may often be computationally prohibitive or restrict the complexity of tables or questions.} TAPAS~\cite{herzig2020tapas} and RCI~\cite{rci} are designed to answer a question by predicting the correct cells in the table in a truly end-to-end manner.  TaBERT \cite{yin2020tabert} is a powerful encoder developed specifically for the TableQA task. 
TaBERT jointly encodes a natural language question and the table, implicitly creating (i)~entity links between question tokens and table-content, and (ii)~relationship between table cells, derived from its structure. To generate the structured query, the encoding obtained from TaBERT is coupled with a memory augmented semantic parsing approach (MAPO) \cite{MAPO}. 


Question generation (QG) \citep{Liu_2020,sultan-etal-2020-importance,shakeri2020endtoend} has been widely explored in reading comprehension (RC) task to reduce the burden of annotating large volumes of Q-A pairs given a context paragraph.  Recently, \citet{puri2020training} used GPT-2~\cite{gpt2} to generate synthetic data for RC, showing that synthetic data alone is sufficient to obtain state-of-art on the SQUAD1.1 dataset. For the QG task in TableQA, systems proposed by \citet{scratchpad2019, sqlQG2018, factoidQG2016} utilize the structure of intermediate logical forms (e.g., SQL) to generate natural language questions. However, none of these QG methods utilize the additional context like table headers, structure and semantics of the tables or the nuances of different possible question types like complex aggregations. To the best of our knowledge, our approach is the first to generate questions specifically for TableQA with the assistance of a logical query and large pre-trained multitask transformers.

Domain adaptation approaches in QA \citep{lee2019domainagnostic, ganin2016domainadversarial} have so far mostly used adversarial learning with an aim to identify domain agnostic features, including in RC applications \cite{wang-etal-2019-adversarial, Cao_Fang_Yu_Zhou_2020}. However, for the TableQA systems using BERT-style language models with vast pre-training, topic shifts remain an unexplored problem. 

\section{T3QA framework}
To our knowledge, this is the first work to explore TableQA in unseen topic setting. Consequently, no public topic-sliced TableQA dataset is available. We introduce a \emph{topic-shift} benchmark by creating new splits in existing popular TableQA datasets: WikiSQL~\cite{zhong2017seq2sql} and Wiki\-TQ~\cite{pasupat2015compositional}. The benchmark creation process is described in Section~\ref{sec:benchmark}. Then, we introduce the proposed framework (illustrated in Figure~\ref{fig:proposed}) to help TableQA system cope with topic shift. Section~\ref{sec:bert_ext} describes the topic specific vocabulary extension for BERT, followed by Question Generation in target topic in Section~\ref{sec:qg} and reranking logical forms in Section ~\ref{sec:our:reranker}.

\subsection{TableQA topic-shift benchmark}
\label{sec:benchmark}

To create a topic-shift TableQA benchmark out of existing datasets, topics have to be assigned to every instance. Once topics are assigned, we create train-test splits with topic shift. I.e., train instances and test instances come from non-overlapping sets of topics. TableQA instances are triplets of the form \{table, question, answer\}. For the datasets WikiSQL and WikiTQ, these tables are taken from Wikipedia articles. WikiSQL has 24,241 tables taken from 15,258 articles and WikiTQ has 2,108 tables from 2,104 articles.

The Wikipedia category graph (WCG) is a dense graph organized in a taxonomy-like structure. For the Wikipedia articles corresponding to tables in WikiSQL and WikiTQ, we found that they are connected to 16000+ categories in WCG on an average. Among the \textit{Wikipedia Category:Main topic articles}, Wikipedia articles were connected to 38+ out of 42 categories in WCG. 

We use category information from Wikipedia articles to identify topics for each of the article and then transfer those topics to the corresponding tables. The main steps are listed below; details can be found in Appendix~\ref{supp:benchmark}.
\begin{itemize}[nosep,leftmargin=*]
\item We identify 42 main Wikipedia categories.
\item For each table, we locate the Wikipedia article containing it.
\item From the page, we follow category ancestor links until we reach one or more main categories.
\item In case of multiple candidates, we choose one based on the traversed path length and the number of paths between the candidate and the article.
\end{itemize}

\noindent
We cannot take an arbitrary subset of topics for train and the rest for test split to create a topic-shift protocol, because many topics are strongly related to others. For example, topic Entertainment is more strongly related to Music than to Law. To avoid this problem, we cluster these Wikipedia main topics into groups such that similar topics fall in the same group.  Using a clustering procedure described in Appendix~\ref{supp:benchmark}, we arrive at 5 high-level topic groups: Politics, Culture, Sports, People and Miscellaneous.

Table~\ref{tab:topics} gives the membership of each topic group and the number of instances in WikiSQL and WikiTQ dataset per topic. For ease of discussion, we will be calling the five topic groups as topics from now on. For both datasets, we create five leave-one-out topic-shift experiment protocols where in each topic becomes the test set, called the \emph{target topic} and the rest four the training set is called the \emph{source topic(s)}. 

In our protocol, for training, apart from the instances from source topic, we also provide tables and document from the target topic. Documents are the text crawled from the target topic articles from Wikipedia. Collecting  unlabeled tables and text data for a target topic is inexpensive. We name these datasets \ws~(WikiSQL with topic shift) and \wtq.

\subsection{Topic specific BERT vocabulary extension}
\label{sec:bert_ext}
Sub word segmentation in BERT has a potential risk of segmenting named entities or in general unseen words in the target corpus. Vocabulary extension ensures that topic specific words are encoded in entirety and avoids splitting into sub-words. Our goal is to finetune BERT with extended vocabulary on topic specific target corpus to learn topic sensitive contextual representation. \eat{In our \ws~ benchmark, we observe that on an average, there is only 1--2\% overlap between target topic vocabulary and BERT's tokenizer vocabulary.} 
So we add frequent topic-specific words to encourage the BERT encoder to learn better topic sensitive representation, which is crucial for better query understanding and query-table entity linking. 
\eat{
We hypothesize that adding frequent topic-specific words will encourage the BERT encoder to learn better topic sensitive representation, which is crucial for better query understanding and query-table entity linking. 
}

\subsection{Table-question generation}
\label{sec:qg}

In our proposed topic-shift experiment protocol with the training set from source topic, unlabeled tables and free text from target topic are provided in the training phase. We propose to use tables from the target topic to generate synthetic question-answer pairs and use these augmented instances for training the TableQA model. Unlike question generation from text, a great deal of additional control is available when generating questions from tables. Similar to \citet{sqlQG2018}, we first sample SQL queries from a given table, and then use a text-to-text transformers (T5)~\cite{raffel2020exploring} based sequence-to-sequence model to transcribe the SQL query to a natural language question. 

\begin{algorithm2e}[!h]
	\scriptsize
	\KwIn{Table T, Integer targetNum}
	\KwOut{SQLQuery[] generatedSQLs} 
     \ForEach {Column C $\in$ T.columns} {
	    $C.dataType \gets ExtractDataType(C)$
	 }
	\While{generatedSQLs.size() $<$ targetNum}{
	    SQLQuery S $\gets$ Empty \;
	    Integer num\_where $\gets$ sample from a multinomial distribution over \{1,2,3,4\} \; \label{line:l1}
	    Where[] whereClauses $\gets$ generateWhereClauses(T, num\_where) \;
        $S.Where$ $\gets$ whereClauses  \;
     	String returnType $\gets$ sample from a multinomial distribution over \{SELECT, SUM, AVG, MAX, MIN\} \; \label{line:l2}
	\If {returnType $\in$ \{SUM, AVG, MAX, MIN\} } { \label{line:l3}
		Column selectColumn $\gets$ sample Column $C$ from \{$T.Columns$ $\backslash$ $whereClauses.Columns$ | $C.dataType=Numeric$\} \;
	} 
	\Else{ 
		Column selectColumn $\gets$ sample Column $C$ from \{$T.Columns$ $\backslash$ $whereClauses.Columns$\} 
	}
    $S.Select$ $\gets$ selectColumn \;
	\If {$\exists$ SQL S2 | $S2.Where \subset S.Where$ AND $S2.Result = S1.result$} {
	    valid = false\; \label{line:l4}
	}
	\If {returnType $\in$ \{SUM, AVG, MAX, MIN\} AND $numOfRows(S.Result)==1$} {
	    valid = false\; \label{line:l5}
	}
	\If{valid} {
	    generatedSQLs.add(S) 
	}
	}
	Return generatedSQLs
	\caption{Algorithm for SQL query Generation from a table. \label{algo:sqlgen}}
\end{algorithm2e}

\subsubsection{SQL sampling}

For generating synthetic SQL queries from a given table~T, we have designed a focused and controllable SQL query generation mechanism presented in Algorithm~\ref{algo:sqlgen}. Our approach is similar to \citet{zhong2017seq2sql} but unlike the existing approaches, we use guidance from target query syntax to offer much more control over the type of natural language questions being generated. We also use additional context such as table header, target answer cell to help the model generate more meaningful questions suitable for \sys. We sample the query type (simple retrieval vs.\ aggregations) and associated where clauses from a distribution that matches the prior probability distribution of training data, if that is available. Sampling of query type and number of where clauses is important to mitigate the risk of learning a biased model that cannot generalize for more complex queries with more than 2 where clauses, as reported by~\citet{sqlQG2018}.

The generated SQL queries are checked for various aspects of semantic quality, beyond mere syntactic correctness in typical rule based generations. \eat{For example, for an aggregation query we check if the field for aggregation is indeed of numeric type.} WikiSQL has a known imitation: even an incorrect SQL query can produce the same answer as the gold SQL query. \eat{Such SQL queries can adversely affect the training pipeline and produce a sub-optimal model.} To avoid such cases, we make two important checks: (1)~The WHERE clauses in the generated SQL queries must all be mandatory to produce the correct answer. i.e.,  dropping a WHERE clause should not produce the expected answer and (2)~a generated SQL query with an aggregation must have at least 2 rows to aggregate on and therefore, dropping the aggregation will not produce the expected answer. These quality checks ensure that the generated synthetic SQL queries are fit to be used in TableQA training pipeline. 

\begin{figure}[!t]
\centering
\includegraphics[width=.48\textwidth]{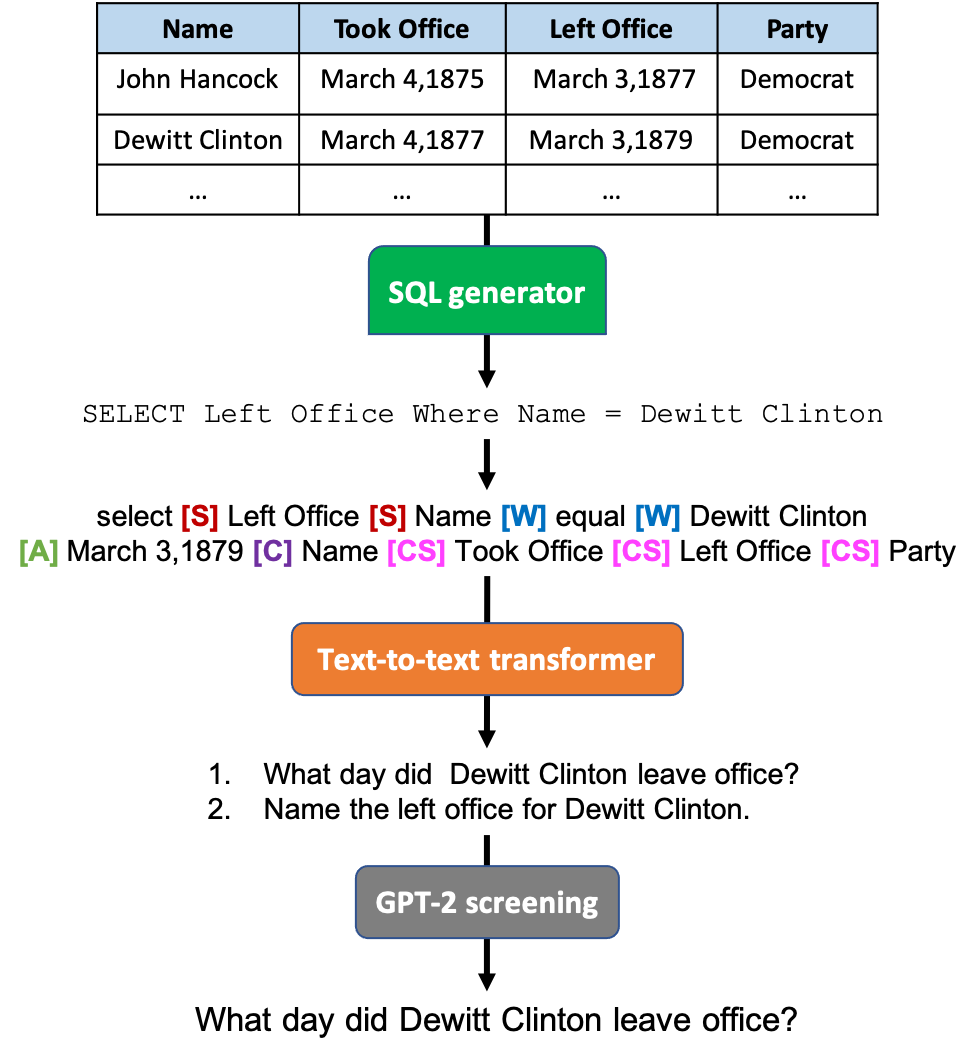}
\caption{Generating synthetic questions on target topics using only tables. Special tokens are shown in colored font.}
\label{fig:t5}
\end{figure}

\begin{table*}[!ht] \footnotesize
\centering
\begin{tabular}{|p{0.07\linewidth}|p{0.27\linewidth}|p{0.27\linewidth}|p{0.27\linewidth}|}
\hline
\textbf{Type} &  \textbf{Ground truth SQL}& \textbf{Generated Question} & \textbf{Ground truth question}  \\ \hline
 & SELECT Rounds WHERE Chassis = b195 & What round has a car with a b195 chassis? &Which rounds had the B195 chassis? \\ 
&&&\\
Lookup &  SELECT College WHERE Player = Paul Seiler & What college does Paul Seiler play for? & What college has Paul Seiler as a player?  \\ 
&&&\\
& SELECT Date WHERE Attendance > 20,066 AND Home = Tampa Bay & On what date was the attendance more than 20,066 at Tampa Bay? & When has an Attendance larger than 20,066 in tampa bay?\\ \hline
& SELECT SUM(Attendance) WHERE Date = May 31 & How many people attended the May 31 game? & How many people attended the game on May 31?\\ 
&&&\\
Aggregate  & SELECT MAX(Mpix) WHERE Aspect Ratio = 2:1 AND Height < 1536 AND Width < 2048 & What is the highest Mpix with an Aspect Ratio of 2:1, a Height smaller than 1536, and a Width smaller than 2048? & What camera has the highest Mpix with an aspect ratio of 2:1, a height less than 1536, and a width smaller than 2048?\\
&&&\\
 & SELECT AVG(Score) WHERE Player = Lee Westwood & What is Lee Westwood's average score? & What is the average score with lee westwood as the player? \\ \hline
\end{tabular}
\caption{\label{tab:t5input} 
Ground truth SQL queries with generated questions (using T5 based QG module) and gold questions}
\end{table*}

\begin{table*}[!t]
\centering  \footnotesize
\begin{tabular}{|p{0.08\linewidth}|p{0.37\linewidth}|p{0.37\linewidth}|}
\hline
\textbf{Operation} &  \textbf{Sampled SQL}& \textbf{Generated Question} \\ \hline
SELECT & SELECT Production code WHERE Written by = José Rivera & what is the production code for the episode written by José rivera? \\ 
& &\\
& SELECT Average WHERE Rank by average > 3 AND Number of dances=17 & what is the average for a rank by average larger than 3 and 17 dances?\\ \hline
MAX & SELECT MAX(SEATS) WHERE Kit/Factory = Factory & can you tell me the highest seats that has the kit/factory of factory? \\ 
& & \\
& SELECT MAX(YEAR) WHERE WINS = 70 AND Manager = Jim Beauchamp & what is the most recent year of the team with 70 wins and manager Jim Beauchamp? \\ \hline
MIN & SELECT MIN(Rank) WHERE Nationality = RUS & which rank is the lowest one that has a nationality ofrus? \\
& & \\
& SELECT MIN(Televote Points) WHERE Panel Points = 0 & which Televote points is the lowest one that has panels pointss of 0? \\ \hline
SUM & SELECT SUM(Game) WHERE Team = Baltimore & what is the sum of game, when team is Baltimore? \\
& & \\
& SELECT SUM(Division) WHERE Year < 2011 AND Playoffs = Did not qualify & what is the total number of division(s), when year is less than 2011, and when playoffs did not qualify? \\ \hline
AVG & SELECT AVG(Digital PSIP) WHERE Network = Omni Television & which digital PSIP has a network of Omni television? \\
& & \\
& SELECT AVG(Attendance) WHERE Week < 5 & what was the average attendance before week 5? \\ \hline

\end{tabular}
\caption{\label{tab:gen_sql_wikisql} Synthetic questions generated on \textit{sampled SQLs} with SELECT and various aggregate functions on \ws tables. Observe that the quality of questions is generally better with SELECT operation than aggregate ones. The reason for this might be that the data used to train QG module includes more SELECT questions.}
\end{table*}

\subsubsection{T5 transfer learning for QG}

For question generation in the TableQA setup, it is more intuitive to create SQL queries first and then use the structure of the SQL query to translate it to a natural language question. Previously, \citet{sqlQG2018} and \citet{scratchpad2019} used LSTM-based sequence to sequence models for direct question generation from tables. However, we hypothesize that apart from SQL queries, using answers and column headers with the help of transformer based models, can be more effective.

For our question generation module we have used unified text-to-text transformers (T5) \cite{raffel2020exploring}, which is popular for its \emph{constrained} text generation capabilities for multiple tasks such as translation and summarization. To leverage this capability of T5 for generating natural language questions from SQL queries, we encode a SQL query in a specific text format. We also pass the answer of the SQL query and the column headers of table to T5 as we observe that using these two sets of extra information along with the SQL query helps in generating better questions, especially with "Wh" words. As illustrated in Figure~\ref{fig:t5}, the generated SQL query with answer and column headers are encoded into a specific sequence before passing onto T5 model. Special separator tokens are used to demarcate different parts of the input sequence: \textbf{[S]} to specify the main column and operation, \textbf{[W] }demarcates elements in a WHERE clause, \textbf{[A]} marks the answer, \textbf{[C]} and \textbf{[CS]} show the beginning of set of column headers and separation between them, respectively. 

In this example, one can observe that although the SQL query do not have any term on day or date, our QG module was able to add ``What day''. Furthermore, ill-formed and unnatural questions generated by T5 model are filtered out using a pretrained GPT-2 model \cite{gpt2}. We removed questions with the highest perplexity scores before passing the rest to the TableQA training module.

For training the QG module, we use SQL queries and questions provided with the WikiSQL dataset. In our experiments, only query+question pairs from the source topics are used to train the question generation module and synthetic questions are generated for the target topic.

We are able to produce high-quality questions using this T5 transcription. Table~\ref{tab:t5input} shows a few example of generated questions from ground truth SQL and Table~\ref{tab:gen_sql_wikisql} on sampled SQLs. Observe that the model is able to generate lookup questions, multiple conditions, and aggregate questions of high quality. It is interesting to see that for the first example in Table~\ref{tab:t5input}, T5 model included the term \emph{car} in the question even though it was not available in the SQL query, probably taking the clue from \emph{chassis}. Some questions created from sampled SQLs for WikiTQ tables is provided in Appendix~\ref{sec:app_qg}.




\subsection{Reranking logical forms}
\label{sec:our:reranker}

We analysed the logical forms predicted by TaBERT model in WikiSQL-TS and observed that the top logical forms often do not have the correct column headers and cell values. In fact, in WikiSQL-TS there is a 15--20\% greater chance of finding a correct prediction from the top-5 predicted logical form than the top~1. 

We propose to use a classifier, Gboost \cite{gboost} to rerank the predicted top-5 logical form. Given a logical form and table-question pair we create a set of features on which a classifier is trained to give higher score to the correct logical form. 


The logical form-question pair which gives the correct prediction is labelled as \emph{+ve} and wrong predictions as \emph{-ve}. We use the predicted logical forms for source topic dev set to train this classifier and in the inference step while predicting for target topic, the logical form which got highest score by the classifier is selected.

\subsubsection{Features for logical form reranker}
Two sets of features are extracted for the reranker: (1)~entity linking based features, (2)~logical form based features.

\noindent \textbf{Entity linking based features:} This captures matches between query fragments and table elements. Our system of entity linking using string matching also finds partial matches. Partial matches happen when only a part of column name or cell value appear in the question. Another scenario is when token in the question partially matches with multiple entities in the table. We create three feature separately for cell values and column headers.
\\
$\bullet$ Number of linked entities in logical form which appear partially or fully in question.\\
$\bullet$ Sum of ratio of tokens matched with entities in logical form. If the questions has word \textit{States} and corresponding entity in table is \textit{United States}, then the ratio would be 0.5.\\
$\bullet$ Sum of a measure of certainty in entity linking. if the question token partially matches with multiple entities in table then certainty is less. If the question has word \textit{United} and there are three entities in the table \textit{United Kingdom}, \textit{United States} and \textit{United Arab Emirates}, then we assign certainty score as 1/3.

\noindent\textbf{Only logical form features:} \\
$\bullet$ Probability score of logical form given by the TableQA model\\
$\bullet$ Length of answer obtained by using this logical form. Length here doesn't mean the number of characters but number of cells in prediction.\\
$\bullet$ If `count' is present in the logical form\\
$\bullet$ If `select' is present in the logical form\\
$\bullet$ Number of where clauses.\\
$\bullet$ If columns are repeated in the logical form.


\section{Experiments and Analysis}
Here we describe key details of the experimental setup, the models compared and evaluation techniques. We also provide a thorough analysis of the results to highlight the key takeaways.

\subsection{Setup}
We consider \ws and \wtq for our experiments with topic assignments as described in Section~\ref{sec:benchmark}. The larger \ws dataset consists of tables, questions and corresponding ground truth SQL queries, whereas \wtq contains only natural language questions and answers. 
The five topics are 1)~\textit{Politics} 2)~\textit{Culture} 3)~\textit{Sports} 4)~\textit{People} and 5)~\textit{Miscellaneous}. Table~\ref{tab:topics} captures some interesting statistics about the topic split benchmark created from WikiSQL. All experiments are conducted in a leave-one-out (LOO) fashion where the target topic examples are withheld. For example, if the target topic is \textit{Politics} then the model is trained using the train set and dev set of \textit{Culture}, \textit{Sports}, \textit{People}, \textit{Misc} and evaluated on test set of \textit{Politics}. Further, a \textbf{composite dev set} is curated by adding equal number of synthetically generated questions from the target topic to the dev set of source topics.

\subsection{Models}
\noindent We perform all experiments using a variant of \base\footnote{\url{https://tinyurl.com/cveejzr6}} ~architecture, with the underlying BERT model initialized with \texttt{bert-base-uncased}. 
\textbf{\base} uses standard BERT as table-question encoder and MAPO~\cite{MAPO} as the base semantic parser. 
\textbf{\tbert} uses topic specific pre-trained BERT encoder (as described in section \ref{sec:bert_ext}). Similar to the base model, this model use MAPO as the base semantic parser. 
\textbf{\baseqg} uses an extended training set with question answer pairs generated from the proposed QG model to train the \base ~model. \textbf{\tbertqg} uses an initialized BERT encoder parameters with topic specific pre-trained BERT and add question-answer pairs generated by our QG model to train the \tbert ~model.
\begin{table}
\resizebox{\hsize}{!}{ \tabcolsep 1pt
\begin{tabular}{|c|c|c|c|c|c|}
\hline
\multicolumn{1}{|l|}{ \textbf{Topic} } & \multicolumn{1}{l|}{ \textbf{TaBERT} } & \multicolumn{1}{l|}{\textbf{ TaBERT$_t$}} & \multicolumn{1}{l|}{\begin{tabular}[c]{@{}l@{}}\textbf{ TaBERT}\\\textbf{+QG} \end{tabular}} & \multicolumn{1}{l|}{\begin{tabular}[c]{@{}l@{}}\textbf{ TaBERT$_t$}\\\textbf{+QG} \end{tabular}} & \multicolumn{1}{l|}{\begin{tabular}[c]{@{}l@{}}\textbf{TaBERT$_t$}\\\textbf{+QG}\\\textbf{+Reranker}\end{tabular}} \\
\hline
Politics & 61.71 & 64.95 & 64.26 & 66.12 & \textbf{70.22} \\
\hline
Culture & 64.89 & 66.10 & 69.32 & 69.88 & \textbf{72.63}  \\
\hline
Sports & 62.10 & 62.70 & 63.03 & 63.83 & \textbf{66.5}  \\
\hline
People & 60.34 & 61.93 & 63.10 & 66.27 & \textbf{70.87} \\
\hline
Misc & 61.85 & 59.03 & 64.31 & 64.43 & \textbf{69.60}  \\
\hline
\end{tabular}
}
\caption{\label{tab:wikisqlres} Performance on \ws benchmark. Here, TaBERT means TaBERT+MAPO and TaBERT$_{t}$ means TaBERT$_{t}$+MAPO. All numbers are in \%.}
\end{table}

\noindent \textbf{Table Question Generation (QG)}: We use the T5 implementation of \citet{huggingface} for question generation, intialized with \texttt{t5-base} and finetuned using SQL and corresponding questions from WikiSQL dataset. To ensure that the target topic is not leaked through the T5 model, we trained five \emph{topic-specific} T5 models, one for each leave-one-out group by considering only SQL-question pairs from the source topic only. As \wtq does not have ground truth SQL queries included in the dataset, we use T5 trained on \ws to generate synthetic questions. We use a batch-size of 10 with a learning rate of~$10^{-3}$.

\noindent \textbf{Implementation details}: We build upon the existing code base for \base~ released by \citet{tabert} and use $\text{BERT}_\text{base}$ as the encoder for tables and questions. 
We use topic-specific vocabulary (explained in Section~\ref{sec:bert_ext}) for BERT's tokenizer and train it using MLM (masked language model) objective for 3 epochs with $p$=0.15 chance of masking a topic-specific high frequency (occurring more than 15 times in target topic corpus) token . We optimize BERT parameters using Adam optimizer with learning rate of $5{\times}10^{-5}$.

All numbers reported are from the test fold, fixing system parameters and model selection with best performance on the corresponding composite dev set. Further details and the dataset are provided in the supplementary material.


\begin{table}\small
\centering
\begin{tabular}{|l|c|c|c|c|}
\hline
\textbf{Topic} & \multicolumn{4}{|c|}{\bfseries Number of WHERE clauses} \\ \hline
 & 1 &   2 &  3 &  4 \\ \hline
Politics &  2.11 &  12.24 &   6.66 &  15.00\\ \hline
Culture  &  0.85 &   8.93 &   4.89 &   5.00 \\ \hline
Sports   &  0.96 &   6.81 &   5.20 &  -3.89  \\ \hline
People   &  1.65 &   9.52 &  11.03 &   6.25 \\ \hline
Misc     &  1.71 &  13.00 &  10.00 & 33.34 \\ \hline
 \end{tabular} 
\caption{\label{tab:wc} Change in performance in \ws{} after applying Reranker to TaBERT$_{t}$+MAPO+QG, across number of WHERE clauses. All numbers are in absolute \%.}
\end{table}

\begin{table*}[!ht] \scriptsize
\centering
\begin{tabular}{|l|l|l|l|l|l|l|l||l|l|l|l|l|l|l|}
\hline
\textbf{\begin{tabular}[l]{@{}c@{}}Topic\end{tabular}} & \multicolumn{7}{c||}{\textbf{\base}} & \multicolumn{7}{c|}{\textbf{\tbertqg+Reranker}} \\ \hline
  \multicolumn{1}{|l|}{} & overall & \multicolumn{1}{c|}{select} & \multicolumn{1}{c|}{count} & \multicolumn{1}{c|}{min} & max & sum & avg & \multicolumn{1}{c|}{overall} & select & count & min & max & sum & avg \\ \hline
Politics  & 61.71 & 62.82& 66.17  &53.28  & 58.64 & 46.26 & 60.21 &70.22  & \textbf{73.90}& 60.59 &70.98  &56.57  &56.71  & 65.59 \\ \hline
Culture  &64.89  &64.47 &  70.62&62.74  &65.56  &62.66  &60.71  & 72.63 & \textbf{74.50} & 65.01 &69.53  &69.93  &64.0  &63.09 \\ \hline
Sports  &62.10  & 61.60 &  57.16  & 69.55 & 72.09 & 54.14 & 62.07 & 66.5 & \textbf{67.06} & 45.45 & 78.85 & 74.39 & 67.15  & 69.41 \\ \hline
People & 60.34 & 59.10&  66.92 &60.71  &69.56  & 50.72 &73.33  & 70.87 & \textbf{72.55}& 60.0 & 72.82 &65.17  &60.86  &  73.33\\ \hline
Misc  &61.85  & 60.8 &  65.0  &72.34  &76.19  &44.82  & 55.17 &69.60  &\textbf{69.76}&66.25 &95.23 \textcolor{green}{$\uparrow$}  &74.46  &44.82  & 51.72 \\ \hline
\end{tabular}
\caption{\label{tab:type_analysis} Performance on WikiSQL Topic specific benchmark across various question types. The largest group, \textit{select}, is shown in \textbf{bold}. Largest improvement is shown as \textcolor{green}{$\uparrow$}. All numbers are in absolute \%.}
\end{table*}



\subsection{Results and Analysis}
\textbf{WikiSQL-TS}: \tbert{} improves over \base{} for four out of five test topics by an average of 1.66\%, showing the advantage of vocabulary extension (Table~\ref{tab:wikisqlres}). In addition to supplying the topic-specific \emph{sense} of vocabulary, fine tuning also avoids introducing word-pieces that adversely affect topic-specific language understanding. For instance, for the topic \textit{culture} the whole word `rockstar' is added to the vocabulary rather than the word-pieces `rocks', `\#\#tar'.
We implement vocabulary extension by using the 1000 placeholders in BERT's vocabulary, accommodating high frequency words from the target topic corpus . 

Further, \baseqg{} sig\-nific\-antly outperforms \base{} and also \tbert{} when finetuned with target topic samples obtained from QG (after careful filtering). In \ws{}, QG also improves the performance of \tbert{}, though relevant vocabulary was already added to BERT, suggesting additional benefits of QG in T3QA framework. While vocabulary extension ensures topical tokens are encoded, QG improves implicit linking between question and table header tokens within the joint encoding of question-table. The largest improvement of 10.53\% and 7.74\% is obtained for People and Culture respectively. Moreover, \baseqg~ out-performs an in-topic performance of 64.07\% and 67\% with 66.27\% and 69.88\% (details in Appendix~\ref{app:app_seen}), showing that the unseen topic performance can be substantially improved with only auxiliary text and tables from documents without explicitly annotated table, question, and answer tuples.

As mentioned, \emph{Misc} is a topic chimera with a mixed individual statistics, hence an explicit injection of frequent vocabulary does not significantly improve \tbert{} over \base. However, \baseqg{} outperforms \tbert{} by 5.4\% due to QG, suggesting that the improvement from both methods are disjoint.
Further, Question generation, though conditioned on the table and topic specific text is not supplied with the topic vocabulary. We also observe that the composite dev set with 50\% real questions and 50\% questions generated on tables from target topic improves performance.
Tables \ref{tab:wikisqlres} \& \ref{tab:wc} take the advantage of ground truth SQL queries to further dissect the performance along question types and number of WHERE clauses.

\begin{table}[!t]
\resizebox{\hsize}{!}{ \tabcolsep 1pt
\begin{tabular}{|c|c|c|c|c|c|}
\hline
\multicolumn{1}{|l|}{ \textbf{Topic} }  & \multicolumn{1}{l|}{ \textbf{TaBERT} } & \multicolumn{1}{l|}{\textbf{ TaBERT$_t$}} & \multicolumn{1}{l|}{\begin{tabular}[c]{@{}l@{}}\textbf{ TaBERT}\\\textbf{+QG} \end{tabular}} & \multicolumn{1}{l|}{\begin{tabular}[c]{@{}l@{}}\textbf{ TaBERT$_t$}\\\textbf{+QG} \end{tabular}} & \multicolumn{1}{l|}{\begin{tabular}[c]{@{}l@{}}\textbf{TaBERT$_t$}\\\textbf{+QG}\\\textbf{+Reranker}\end{tabular}} \\
\hline
Politics & 40.52 & 41.03 & 41.55 & 41.38 & \textbf{43.79} \\ \hline
Culture & 36.03 & 38.49 & 38.49 & 37.05 & \textbf{39.50}  \\ \hline
Sports & 37.55 & 37.5 & 37.93 & 39.12 & \textbf{41.50}   \\ \hline
People & 35.94 & 37.69 & 37.42 & 36.61 & \textbf{39.30} \\ \hline
Misc & 38.58 & 40.64 & 41.10  & 40.18 & \textbf{42.23} \\ \hline
\end{tabular}
}
\caption{\label{tab:wtqres} Performance on WikiTQ-TS benchmark. Here, TaBERT means TaBERT+MAPO and TaBERT$_{t}$ means TaBERT$_{t}$+MAPO. All numbers are in \%.}
\end{table}

\noindent \textbf{Number of Where clauses:} As described previously, performance of \base ~is substantially affected by the number of WHERE clauses in the ground truth logical form (also observed by \cite{sqlQG2018}), see Appendix~\ref{sec:app_wikisql}. Table~\ref{tab:wc}, 
shows that performance improvement by ``Reranker" is significantly higher for more than 1 WHERE clause. This might have happened because \base~ prefers to decode shorter logical forms, whereas the reranker prioritizes logical forms with more linked entities present from the question.



\noindent \textbf{WikiSQL question types}: Table \ref{tab:wikisqlres} breaks down the performance of \baseqg~ based on the question types labels obtained from the dataset ground truth only for analysis. The improvement, viewed from the lens of question types is more significant with average gain in SELECT-style queries at 9.76\%. Aggregate (count, min/max, sum, avg) questions are more challenging to generate as the answer is not present in the table. Consequently, the performance improvement with QG is less significant for these question types.


\noindent \textbf{WikiTQ-TS}: WikiTQ-TS is a smaller dataset and contains more complex questions (negatives, implicit nested query) compared to WikiSQL-TS. Correspondingly, there is also less topic specific text to pretrain the TaBERT encoder. Despite these limitations, we observe in Table~\ref{tab:wtqres} that TaBERT$_t$ with vocabulary extension and pretraining shows overall improvement. We resort to using synthetic questions generated from QG model of \ws, due to unavailability of ground truth SQL queries in WikiTQ.
Hence, the generated questions are often different in structure from the ground truth questions. Samples of real and generated questions are in Table~\ref{tab:app_qg_wtq} of Appendix~\ref{sec:app_qg}. Despite this difference in question distribution we see TaBERT+QG consistently performs better than the baseline. We provide an analysis of the perplexity scores from TaBERT and TaBERT$_t$ on the generated questions in Appendix~\ref{sec:prep_qg}. Ultimately, the proposed T3QA framework significantly improves performance in all target domains.

\section{Conclusion}
This paper introduces the problem of TableQA for unseen topics. We propose novel topic split benchmarks over WikiSQL and WikiTQ and highlight the drop in performance of TaBERT+MAPO, even when TaBERT is pretrained on a large open domain corpora. We show that significant gains in performance can be achieved by (i)~extending the vocabulary of BERT with topic-specific tokens (ii)~fine-tuning the model with our proposed constrained question generation which transcribes SQL into natural language, (iii)~re-ranking logical forms based on features associated with entity linking and logical form structure. We believe that the proposed benchmark can be used by the community for building and evaluating robust TableQA models for practical settings.

\bibliography{tableqa}
\bibliographystyle{acl_natbib}

\newpage
\clearpage
\newpage

\appendix

\twocolumn[\centering\Large\bfseries\ztitle\\
(Appendix) \par \medskip]

\begin{table*}[]
\centering  \footnotesize
\begin{tabular}{|p{0.45\linewidth}|p{0.45\linewidth}|}
\hline
\textbf{Gold questions in the dataset }& \textbf{Generated Questions} \\ \hline
- how many v8 engines competed in the 1982 British formula one season? & - which constructor has an Entrant of Colin Bennett racing, and a no smaller than 7.0?\\
- how many entrants have names that contain the word "team"? & - what is the average number that has a constructor of Ensign?\\
- name an entrant with no cosworth engines. & - who is the driver with a no of 14.0?\\
- how many drivers use v8 engines? & - what is the average number of FW07 chassisassis?\\
- what is the total number of drivers listed? & - what engine has a driver of Steveo'rourke?\\
- who is the only driver to use a v12 engine? & - name the most number for Chassis being N180b.\\
- Are there any other engines listed besides cosworth or brm? & - what is the lowest number that has a constructor of Ensign?\\
- Which is the only driver whose vehicle used a brm 202 v12 engine? & - what is the total number that has a constructor of Williams?\\
- What is the last chassis listed? & - what is the largest number for teamensign?\\ \hline
\end{tabular}
\caption{\label{tab:app_qg_wtq} Real questions and generated questions for table \texttt{csv/203-csv/1.tsv} from WikiTQ-TS dataset. Observe that generated questions (finetuned on \ws) have different semantics than the real questions.}
\end{table*}

\section{TaBERT performance on WikiSQL}
\label{sec:app_wikisql}

\begin{figure}[!ht]
\centering
\subfigure[]{\includegraphics[ width=0.5\linewidth]{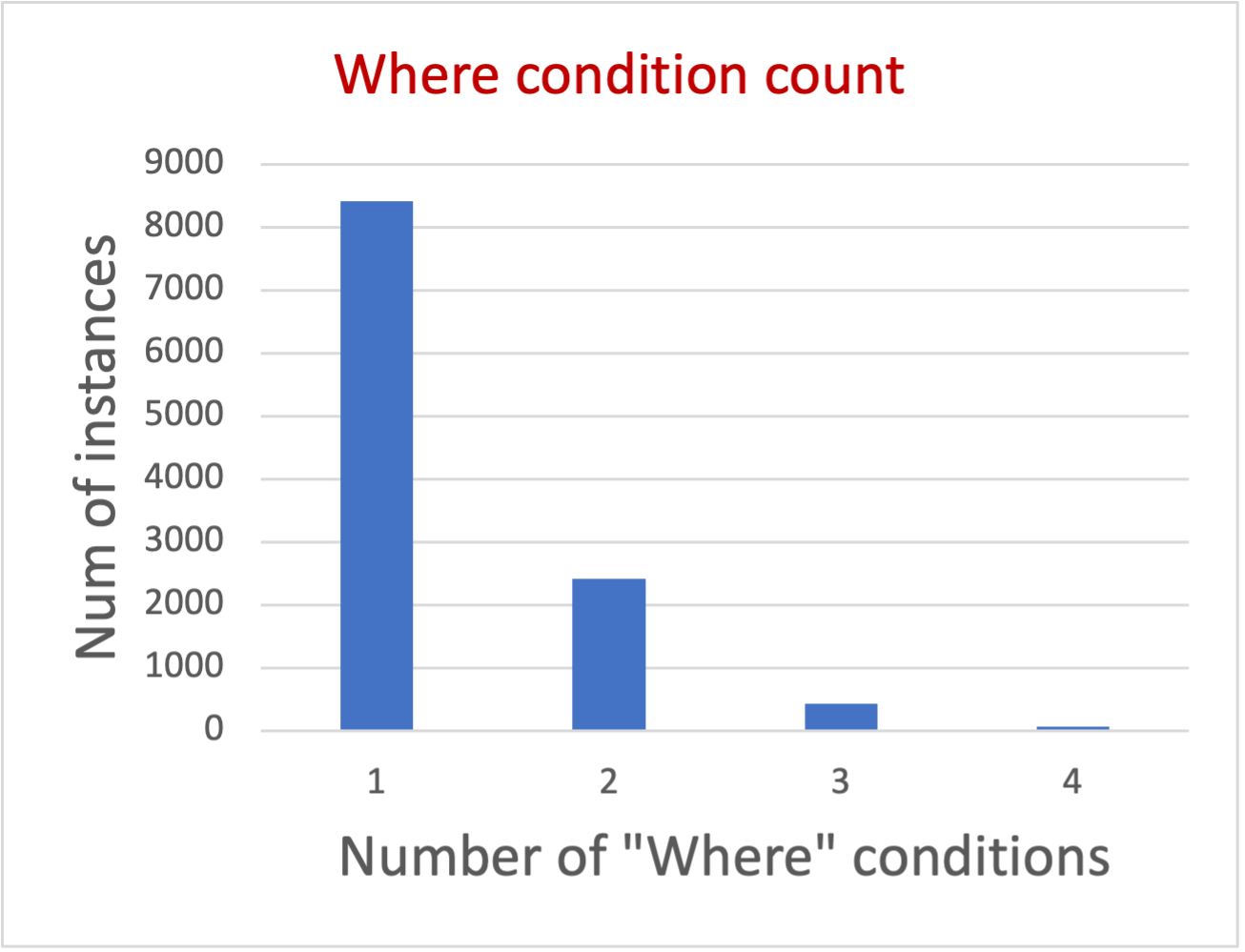}}\hfill
\subfigure[]{\includegraphics[width=0.5\linewidth]{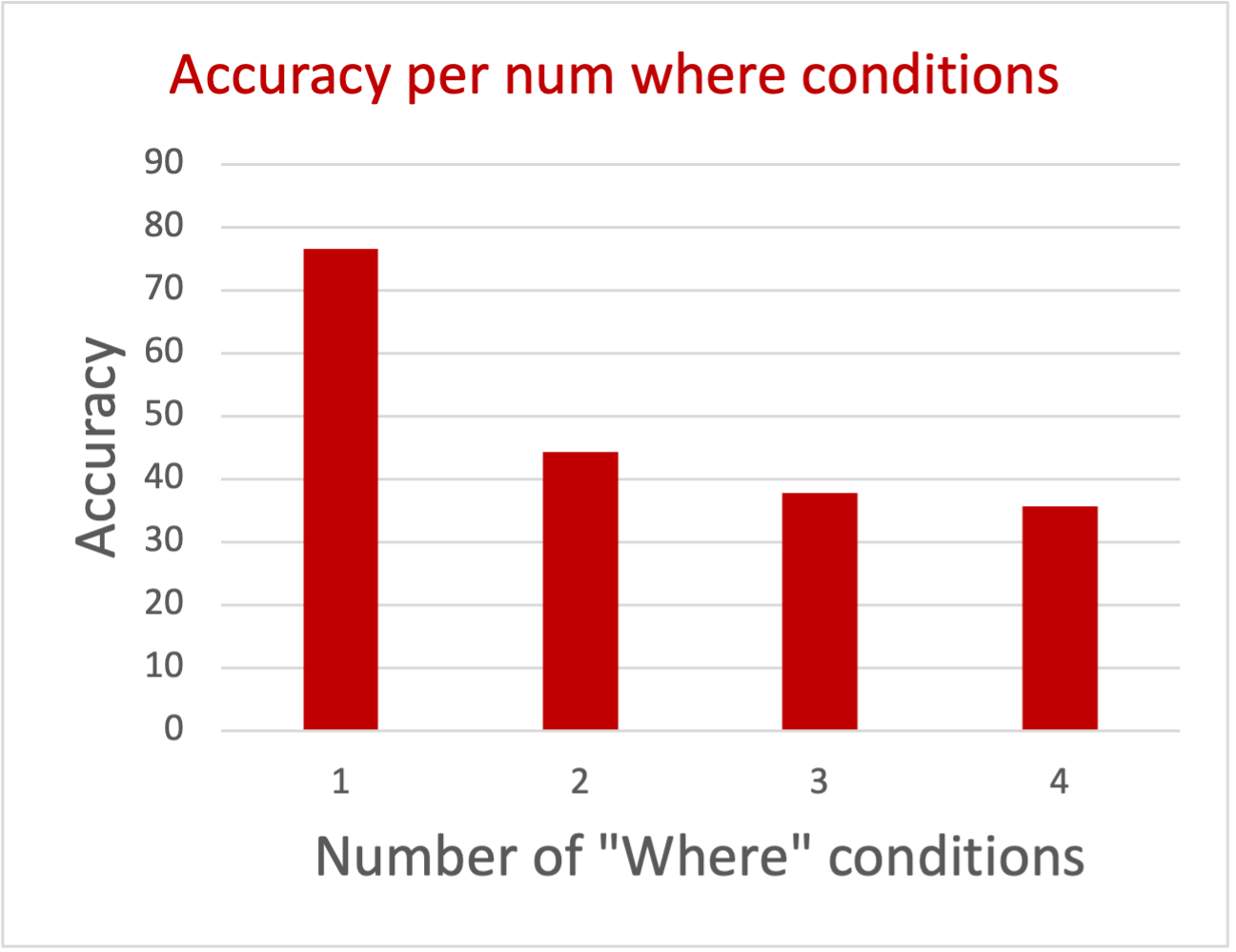}}\hfill\\
\subfigure[]{\includegraphics[ width=0.5\linewidth]{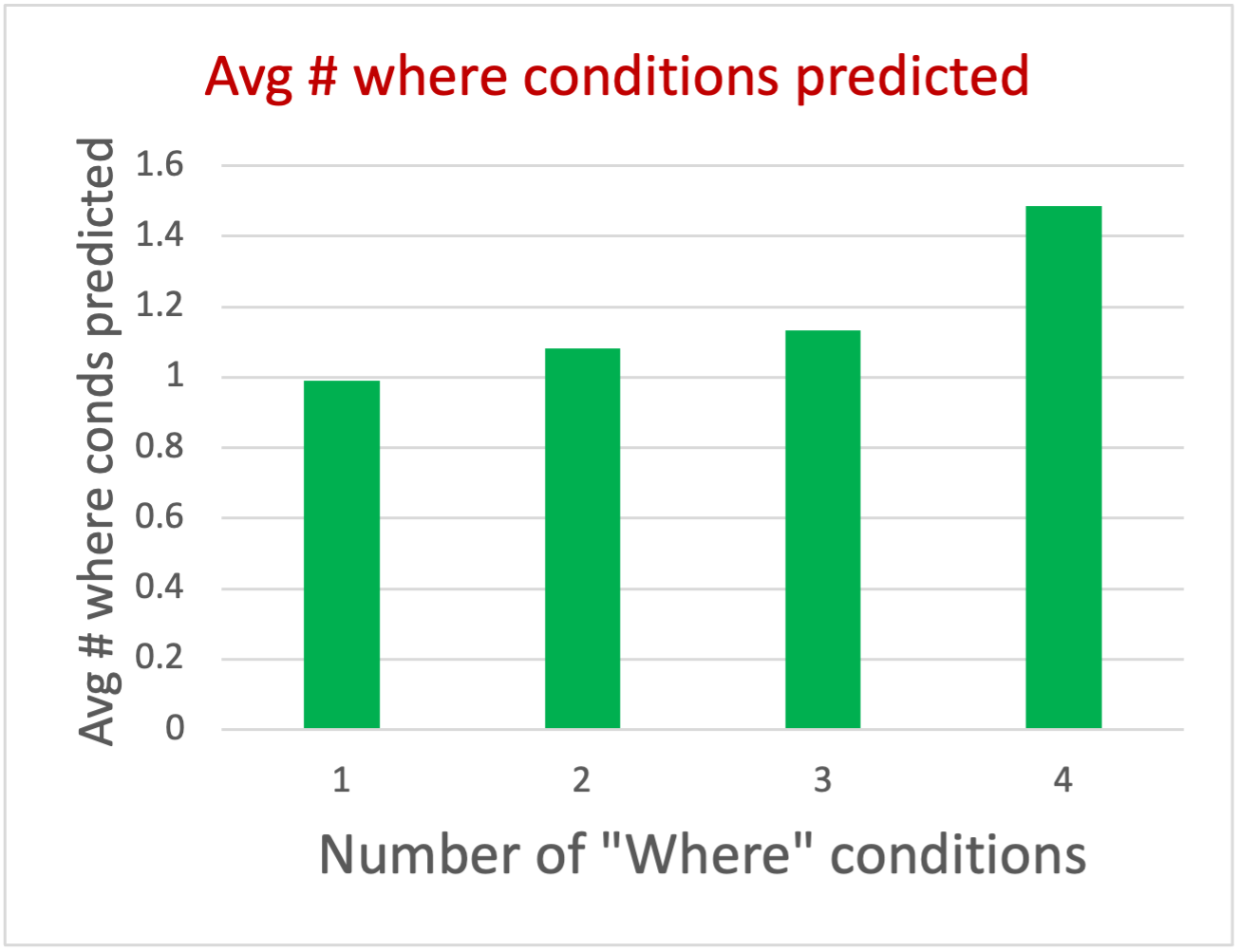}}\hfill
\subfigure[]{\includegraphics[ width=0.5\linewidth]{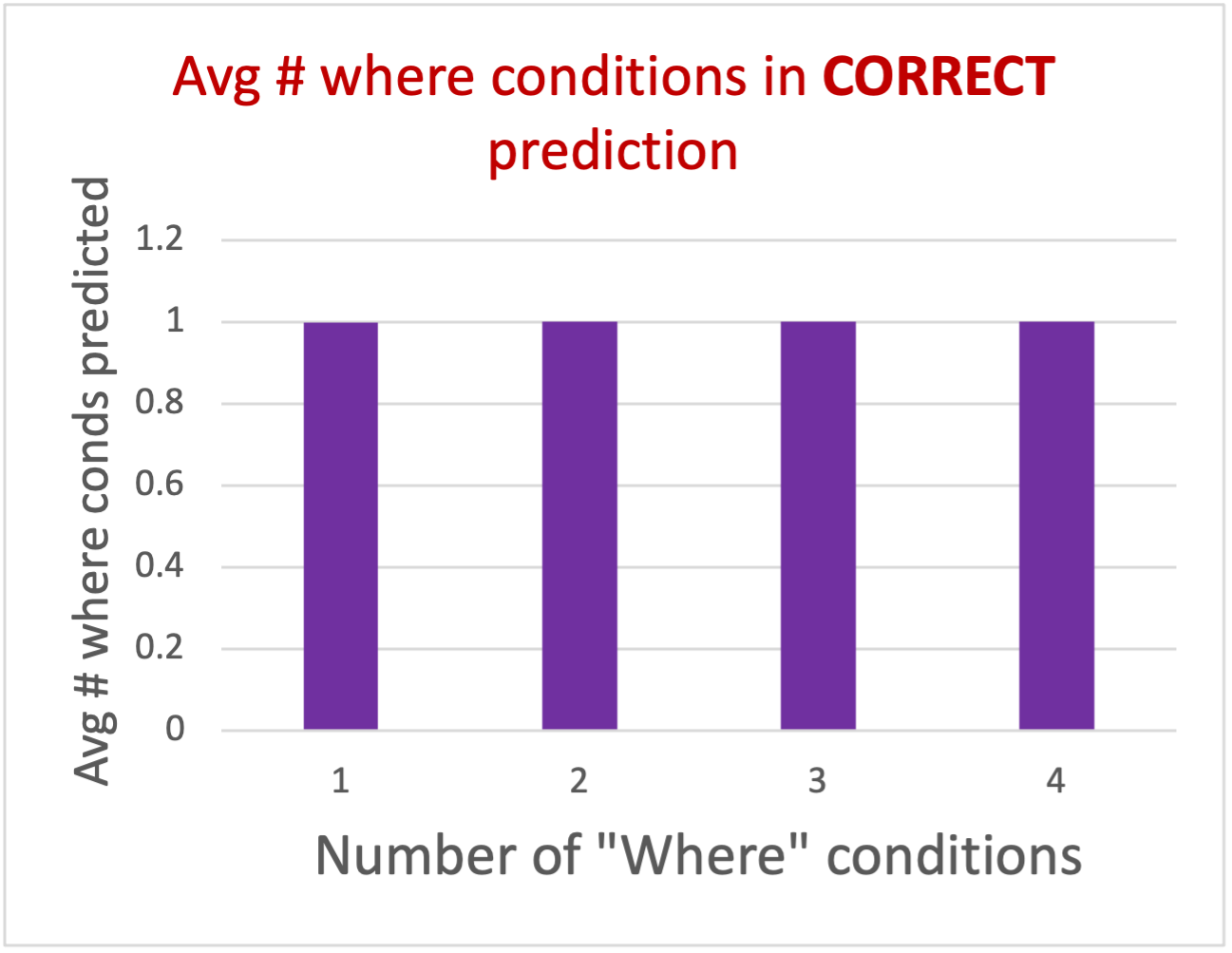}}
\caption{WHERE clause analysis on TaBERT+MAPO performance on SELECT questions in WikiSQL test set (not topic shift): (a)~Frequency of questions in different  number of WHERE clauses buckets; (b)~Accuracy achieved in each WHERE bucket; (c)~Average number of conditions in predicted logical forms; (d)~Average number of conditions in predicted logical forms which produces correct answers.}
\label{fig:where}
\end{figure}

We analyse accuracy of TaBERT model on Wiki\-SQL in terms of the number of WHERE clauses, which are skewed as shown in Fig.~\ref{fig:where}(a). In Fig.~\ref{fig:where}(b), we observe that accuracy decreases when ground truth SQL has a larger number of WHERE clauses. Interestingly, we observe in Fig.~\ref{fig:where}(c) and (d) that even though the model achieves 30\% to 40\% accuracy for 2--4 WHERE clauses, the predicted logical form still produced one WHERE clause. This shows that, for many questions, wrong or incomplete logical forms can produce correct answers.

\section{Topic-shift benchmark details}
\label{supp:benchmark}
Continuing from Section~\ref{sec:benchmark}, this section provides more details about the creation of the topic shift benchmark datasets.
Each Wikipedia article is tagged with a set of categories and each category is further tagged with a set of parent categories, and those to their parent categories, and so on. The whole set of Wikipedia categories are organized in a taxonomy-like structure called Wikipedia Category Graph (WCG)~\cite{wcg}. These categories range from specific topics such as "Major League Soccer awards" to general topics such as "Human Nature". To have categories of similar granularity, we use the 42 categories listed in \textit{Wikipedia Category:Main topic articles}\footnote{\url{https://w.wiki/458k}} as topics.

To assign a unique category to a Wikipedia article, we proceed as follows:
\eat{For a given article, the topic with the shortest path to the article in the Wikipedia Category Graph is assigned. In case of tie, when multiple topics are at the same distance to the article, number of different paths topic node can reach the article node is counted and the one with the most of paths is assigned. After doing this exercise, we get every table assigned to one of these 42 topics.
}

\noindent$\bullet$ For each Table $T$, we extract the Wikipedia Article $A$ which contains Table $T$.  \\
$\bullet$  We start with the category of $A$ and traverse the hierarchical categories till we reach one (or more) of the 42 categories listed in \textit{Wikipedia Category:Main topic articles}. \\
$\bullet$  If multiple main topic categories can be reached from $A$, we take the category which is reached via the shortest path (in terms of number of hierarchical categories traversed from $A$) and assign that as the category for table $T$. \\
$\bullet$  If there are multiple main topic categories which can be reached with the same length of shortest path, we consider the number of different paths between the main topic category and $A$ as the tie breaker to assign the topic for $A$.

Now we describe the method used to cluster categories into topics.
For every article we identify five categories closest to the article in Wikipedia Category Graph. We then compute the Jaccard similarity between two topics as the ratio of number of common articles between topics (in the first-5 list) to the total number of articles assigned to both topic. Using this similarity, we apply spectral co-clustering \cite{dhillon2001co} to form five topic groups.

To verify the coherence of the five topic groups, we performed a vocab overlap exercise. For questions in WikiTQ, we find the 100 most frequent words in the test set of each of the topics. Then we measure how many of these frequent words appeared in the train set of each of these topics. Table~\ref{tab:vocab_match} shows the that word overlap is large within clusters.

\begin{table}[!ht] \scriptsize
\resizebox{\hsize}{!}{%
\begin{tabular}{|l|c|c|c|c|c|}
\hline
Test/Train & Politics & Culture & Sports & People & Misc \\ \hline
Politics   & \textbf{88}& 73 & 74 & 72& 64\\ \hline
Culture    & 79  & \textbf{87} & 89& 85& 69\\ \hline
Sports     & 67  & 72 & \textbf{100} & 81& 50\\ \hline
People     & 66  & 78 & 88 & \textbf{93} & 55\\ \hline
Misc       & 74  & 73 & 74& 72& \textbf{68} \\ \hline
\end{tabular}
}
\caption{Percent vocabulary match within and across topics (category groups/clusters).}
\label{tab:vocab_match}
\end{table}

\section{Questions generation for target topics}
\label{sec:app_qg}

Table~\ref{tab:app_qg_wtq} compares ground truth questions with that of generated questions for the same table from \ws. One can see that even template of questions in real dataset is very different and often tougher than the generated ones. Question generator being trained on \ws dataset with much simple questions might be the reason for this phenomenon.

\section{Performance when topics are seen}
\label{app:app_seen}
We further analyse the performance of the model in both seen-topic training (when the topic specific train set is available), against the unseen topic train (when the topic specific train set is not used during training). In Table~\ref{tab:domain_test}, we present results in both training setups.

\begin{table}[ht]
\small
    \centering
    \begin{tabular}{|c|c|c|}
    \hline
    \textbf{Topic} & \textbf{Seen Topic} & \textbf{Unseen Topic}\\
    \hline
       Politics  & \textbf{65.52} & 61.71 \\
       Culture & \textbf{67.26} & 64.88 \\
       Sports & \textbf{63.14} & 62.10 \\
       People & \textbf{64.07} & 60.34 \\
       Misc & \textbf{63.14} & 61.85 \\
       \hline
    \end{tabular}
    \caption{Drop in performance due to topic shift in \ws.  (Numbers are percentages.)}
    \label{tab:domain_test}
\end{table}


\section{Additional Experiments}
Table \ref{tab:wc2} shows the absolute values corresponding to Table 6. in the paper. The performance of both models is lower for questions with larger WHERE clauses. Table \ref{tab:wc3} summarizes the answer accuracy of TaBERT$_{t}$+MAPO +QG +Reranker and TaBERT+MAPO across number of where clauses in the ground truth logical forms.
\begin{table}[ht!]
\resizebox{\hsize}{!}{%
\tabcolsep 1pt
\begin{tabular}{|l|c|c|c|c|}
\hline
Topic & \multicolumn{4}{|c|}{\bfseries Number of WHERE clauses} \\ \hline
 & 1 &   2 &  3 &  4 \\ \hline
Politics &  75.78/73.67 &  58.36/46.12 &   51.66/45.0 &    40.0/25.0\\ \hline
Culture  &  77.52/76.67 &  61.23/52.30 &  52.44/47.55 &    55.0/50.0 \\ \hline
Sports   &  71.26/70.30 &  57.62/50.81 &   56.26/51.06 &  48.05/51.94 \textcolor{red}{$\downarrow$} \\ \hline
People   &  77.61/75.96 &  61.37/51.85 &  58.82/47.79 &   25.0/18.75 \\ \hline
Misc     &  75.17/73.46 &    57.0/44.0 &    54.0/44.0 &  66.67/33.33\\ \hline
 \end{tabular} }
\caption{\label{tab:wc2} \ws~ performance for TaBERT$_t$+MAPO +QG+Reranker and TaBERT$_t$+MAPO+QG (seperated by `/') across number of WHERE clauses in the ground truth logical forms. All numbers are in \%.}
\end{table}

\begin{table}[ht!]
\resizebox{\hsize}{!}{%
\tabcolsep 1pt
\begin{tabular}{|l|c|c|c|c|}
\hline
Topic & \multicolumn{4}{|c|}{\bfseries Number of WHERE clauses} \\ \hline
 & 1 &   2 &  3 &  4 \\ \hline
Politics &  75.78/67.61 &  58.36/46.94 &   51.66/48.33 &    40.0/40.0\\ \hline
Culture  &  77.52/71.36 &  61.23/46.46 &  52.44/51.05 &    55.0/35.0 \\ \hline
Sports   &  71.26/67.78 &  57.62/50.66 &   56.26/53.9 &  48.05/44.16 \\ \hline
People   &  77.61/69.55 &  61.37/44.62 &  58.82/45.59 &   25.0/37.5\textcolor{red}{$\downarrow$} \\ \hline
Misc     &  75.17/70.58 &    57.0/40.5 &    54.0/46.0 &  66.67/66.67\\ \hline
 \end{tabular} }
\caption{WikiSQL-TS performance for TaBERT$_{t}$+MAPO +QG +Reranker and TaBERT+MAPO (separated by `/') across number of WHERE clauses in the ground truth logical forms.}
\label{tab:wc3}
\end{table}

\section{Training details}
\label{sec:train_details}
We train all \base~ variants for 10 epochs on 4 Tesla V100 GPUs using mixed precision training\footnote{\url{https://github.com/NVIDIA/apex}}. For training \base~, we set batch size to 10, number of explore samples 10 and other hyperparameters are kept same as \cite{yin2020tabert}. We build upon codebase\footnote{\url{https://github.com/pcyin/pytorch\_neural\_symbolic\_machines}} released by \cite{yin2020tabert}. The hyper-parameters (where not mentioned explicitly) are the same are the original code. We include all the data splits and predictions from our best model as supplementary material with the paper. These will be released publicly upon acceptance. The experimentation requires for 5 topics, we performed 6 variations of the model. We performed search over 4 sets of hyper-parameters, primarily on the composition of generated vs. real questions.

\section{TaBERT vs. TaBERT$_{t}$ perplexity of generated questions for WikiTQ-TS}
\label{sec:prep_qg}
We compute the perplexity scores over a subset of 50 generated questions used in the experiments using both TaBERT and TaBERT$_{t}$ language models. Note that TaBERT is pretrained on large open domain set whereas TaBERT$_{t}$
was further fine-tuned on topic specific documents closely related to the tables of target domain. As shown in Table \ref{tab:perpscore}, the average perplexity score from TaBERT$_{t}$ is larger than  TaBERT. This indicates that the generated questions are not aligned to the topic in the case of WikiTQ-TS. This is due to the lack of any training examples for specific to the dataset, as mentioned in Section 4.3. Future work on topic-specific question generation may address this issue.

\begin{table}[!h]
\centering
\begin{tabular}{|l|c|c|}
\hline
Topic    & TaBERT & TaBERT$_{t}$ \\ \hline
Politics & 1.088 & 1.112 \\ \hline
Culture  & 1.099 & 1.142 \\ \hline
Sports   & 1.084 & 1.134 \\ \hline
People   & 1.109 & 1.164 \\ \hline
Misc     & 1.104 & 1.153 \\ \hline
\end{tabular}
\caption{\label{tab:perpscore} The average perplexity scores of a subset of generated questions from TaBERT and TaBERT$_{t}$ for WikiTQ-TS}
\end{table}
We suspect that this might be the reason why TaBERT$_t$+QG does not outperform TaBERT+QG in the case of WikiTQ-TS (Table~\ref{tab:wtqres}). However, we obtain best performance via the overall T3QA framework.

\clearpage



\end{document}